\documentclass[10pt,twocolumn,letterpaper]{article}

\usepackage[accsupp]{axessibility}
\usepackage{iccv}
\usepackage{times}
\usepackage{epsfig}
\usepackage{graphicx}
\usepackage{amsmath}
\usepackage{amssymb}
\usepackage{authblk}

\usepackage[pagebackref=true,breaklinks=true,letterpaper=true,colorlinks,bookmarks=false]{hyperref}

\iccvfinalcopy 

\def\iccvPaperID{3118} 

\ificcvfinal\fi

\begin{document}

\title{Track without Appearance: Learn Box and Tracklet Embedding with Local and Global Motion Patterns for Vehicle Tracking}

\author[1]{Gaoang Wang}
\author[2]{Renshu Gu}
\author[1]{Zuozhu Liu}
\author[3]{Weijie Hu}
\author[1]{Mingli Song}
\author[4]{Jenq-Neng Hwang}
{\affil[1]{Zhejiang University, }}
{\affil[2]{Hangzhou Dianzi University, }}
{\affil[3]{Guangdong University of Petrochemical Technology, }}
\affil[4]{University of Washington}

\affil[ ]{{\tt\small \{gaoangwang, zuozhuliu\}@intl.zju.edu.cn, renshugu@hdu.edu.cn, huweijie@gdupt.edu.cn, brooksong@zju.edu.cn, hwang@uw.edu}}

\maketitle
\ificcvfinal\thispagestyle{empty}\fi

\begin{abstract}
   Vehicle tracking is an essential task in the multi-object tracking (MOT) field. A distinct characteristic in vehicle tracking is that the trajectories of vehicles are fairly smooth in both the world coordinate and the image coordinate. Hence, models that capture motion consistencies are of high necessity. However, tracking with the standalone motion-based trackers is quite challenging because targets could get lost easily due to limited information, detection error and occlusion. Leveraging appearance information to assist object re-identification could resolve this challenge to some extent. However, doing so requires extra computation while appearance information is sensitive to occlusion as well. 
   In this paper, we try to explore the significance of motion patterns for vehicle tracking without appearance information. We propose a novel approach that tackles the association issue for long-term tracking with the exclusive fully-exploited motion information. We address the tracklet embedding issue with the proposed reconstruct-to-embed strategy based on deep graph convolutional neural networks (GCN). 
   Comprehensive experiments on the KITTI-car tracking dataset and UA-Detrac dataset show that the proposed method, though without appearance information, could achieve competitive performance with the state-of-the-art (SOTA) trackers. The source code will be available at \url{https://github.com/GaoangW/LGMTracker}.
\end{abstract}

\section{Introduction}

\begin{figure}[t]
\begin{center}
\includegraphics[width=0.9\linewidth]{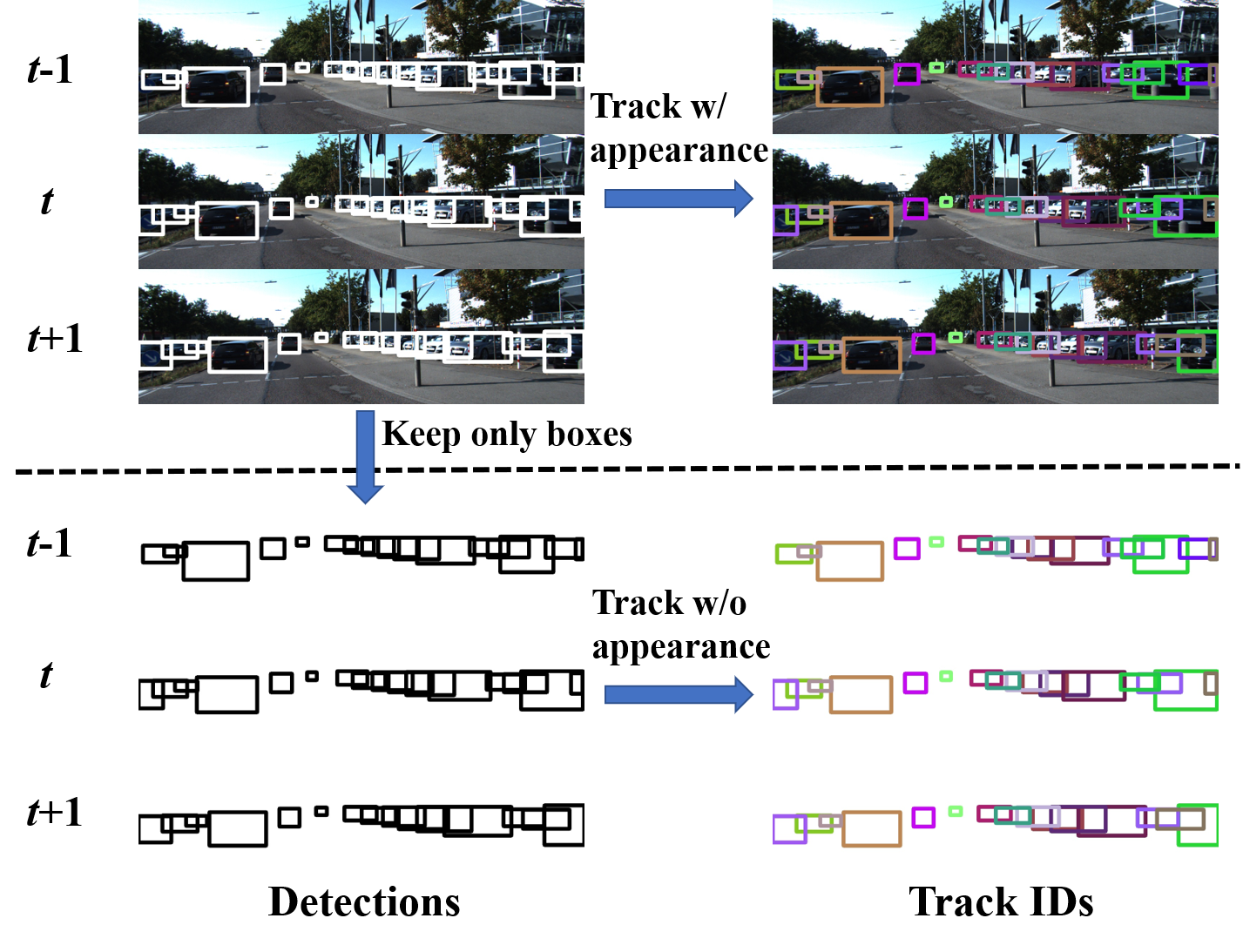}
\end{center}
   \caption{The top part shows tracking with appearance information, while the bottom shows tracking using detection boxes without employing appearance. Obviously, it is more challenging for tracking only based on motion information.}
\label{fig:det_vis}
\end{figure}

Multi-object tracking (MOT) is an important topic in the computer vision and machine learning field. This technique is highly demanded in many tasks, such as traffic flow estimation, human behavior prediction and autonomous driving assistance \cite{tang2018single,tang2019cityflow,wang2019anomaly,hsu2019multi,gu2019efficient}. From unsupervised rule-based \cite{1517Bochinski2017,1547Bochinski2018,bewley2016simple,gunduz2019efficient,tang2018single} and optimization-based \cite{zhang2008global,choi2015near,xiang2015learning,wang2017learning,lenz2015followme,milan2013detection,milan2013continuous,andriyenko2012discrete,hornakova2020lifted,chu2013tracking} to deep learning-based trackers \cite{chu2019famnet,zhou2020tracking,peng2020chained,wang2020joint,pang2020tubetk,bergmann2019tracking,zhang2020long,peng2020tpm,xu2020train}, significant progress of the MOT techniques has been made in the recent ten years. However, some critical challenges still remain. For example, occlusion is still one of the major issues. Without occlusion handling, the targets can easily get lost and identities may get switched. Other challenges, such as crowded scenarios, detection errors and camera motions, also have significant influences on a tracker's performance.

Appearance information is widely used for MOT and greatly improves performance. Appearance information is employed either in an association manner \cite{tang2018single,zhang2020long,chuang2014tracking} or with regression-based approaches for joint learning of detection and tracking \cite{zhou2020tracking,peng2020chained,wang2020joint,bergmann2019tracking}. The assumption, as well as the attribution of its success, is that the same targets from adjacent frames should share similar appearance features. However, the appearance feature is still sensitive to occlusions and objects may have quite different appearance representations when they are occluded. Additionally, joint learning approaches require an extra computational cost.  

Motion consistency is another cue that can be taken advantage of for MOT, especially for vehicle tracking scenarios. This is based on the assumption that the motions of objects usually follow fairly smooth patterns in both the world coordinate and the image coordinate. In particular, for objects that cannot change the orientation and speed rapidly, such as vehicles, motion consistency could play a pivot role for tracking. In addition, the motion feature, usually with the four bounding box parameters for each object, is simple and light, saving more computations than complex appearance features. As a result, mere motion trackers are still worth exploring.
However, there are two main difficulties to establish deep motion-based models.
First, motion itself can only provide limited information. As shown in Figure~\ref{fig:det_vis}, after discarding all appearance information, it is highly challenging to associate the bounding boxes correctly even for humans when false positives and false negatives occur in the detection.
Second, to alleviate the long-term occlusion issue, tracklet association is needed in deep motion-based models. Accurate association requires expressive tracklet embeddings that could be used to measure the similarity among different tracklets. However, learning such embeddings is very challenging as we need to capture temporal consistency as well. For example, tracklets of the same object might have different temporal lengths or do not share similar locations along time, leading to inferior embeddings in practice. 
Due to the aforementioned challenges, mere motion trackers usually cannot achieve comparable performance with models that adopt appearance information. 

\begin{figure*}[t]
\begin{center}
\includegraphics[width=0.9\linewidth]{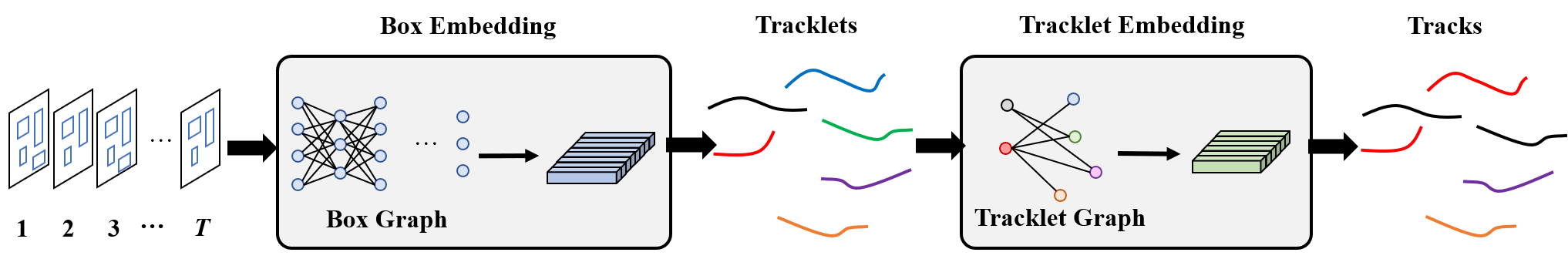}
\end{center}
   \caption{The flowchart of the local-global motion (LGM) tracker. The LGM tracker contains two modules, the box embedding module and the tracklet embedding module, which exploit the local and global motion patterns, respectively. Specifically, the box embedding module embeds input boxes and connects them into tracklets, and the tracklet embedding module aims to associate tracklets into tracks. Both modules are learned based on deep graph convolutional networks (GCN).}
\label{fig:Flowchart}
\end{figure*}

In this paper, we tackle the vehicle tracking problem only from the motion perspective. Without appearance information, we aim to explore how well a motion-based model can perform for the vehicle tracking task. A novel motion-based tracking approach, i.e., local-global motion (LGM) tracker, is proposed to exploit the motion consistency without using any appearance information after the detection. More specifically, without appearance information means: 1) NO further bounding box regression or refinement from the detector feature maps; 2) NO appearance information used for further association and re-identification. 
The flowchart of the proposed LGM tracker is shown in Figure~\ref{fig:Flowchart}. We model the MOT problem as a two-stage embedding task where both local and global motion consistencies are utilized. At the first stage, we aim at learning the box embedding based on deep graph convolutional neural networks (GCN) to associate boxes into tracklets. Since such local associations cannot capture the global track patterns, the occlusion issue is yet unaddressed. 
To break through such limitation, at the second stage, the tracklet embedding with global motion consistency is learned to further associate tracklets into tracks. To better model the tracklet embedding, a novel embedding strategy, \textit{reconstruct-to-embed}, is proposed with the temporal gated convolution mechanism under an attention-based GCN. 

Our contributions are summarized as follows:
1) we tackle the vehicle tracking task from the motion perspective without using appearance information;
2) we propose a novel box and tracklet embedding method that can utilize both the local and global motion consistencies;
3) we evaluate the proposed LGM tracker on KITTI \cite{Geiger2012CVPR} and UA-Detrac \cite{wen2020ua} benchmark datasets and achieve competitive performance with the state-of-the-art (SOTA) trackers.


\section{Related Work}

\subsection{Motion Models}
Motion trackers \cite{1517Bochinski2017,sanchez2016online,hong2016online,yoon2015bayesian,babaee2018occlusion,fang2018recurrent} without using appearance information are also studied in recent a few years. For such methods, some use fairly simple rules, like intersection-over-union (IOU) between adjacent frames as association \cite{1517Bochinski2017}; some adopt particle filter framework in the tracking \cite{sanchez2016online}; some apply recurrent neural networks (RNN) to learn the motion patterns \cite{fang2018recurrent,babaee2018occlusion}. However, with limited information, good performance cannot be easily achieved by motion-based methods. 


\subsection{Graph Models}
Conventional graph models \cite{tang2017multiple,milan2016multi,tang2018single,keuper2016multi,kumar2014multiple,choi2015near,tang2015subgraph,wen2014multiple,wang2019exploit,wang2017learning,lenz2015followme,milan2013continuous,andriyenko2012discrete,hornakova2020lifted} are widely used in MOT for data association. Usually, detections or tracklets are adopted as graph nodes. Then the similarities among nodes are measured on the connected edges. The association is solved by optimizing the total cost or energy function.
However, the similarity measure is usually based on hand-crafted feature fusion, requiring empirically setting a lot of hyper-parameters.
Since graph neural networks (GNN) show great power recently, many approaches \cite{braso2020learning,shan2020fgagt,weng2020gnn3dmot} adopt GNN for the association, rather than using the conventional graph models based on optimization. However, for most existing methods based on GNN, only the local association based on adjacent frames is considered. As a result, the long-term occlusion is still one major issue for GNN based trackers.

\subsection{Joint Detection and Tracking}
More recently, joint detection and tracking based methods have been drawn great attention \cite{zhou2020tracking,peng2020chained,wang2020joint,pang2020tubetk,bergmann2019tracking}. Usually, the tracker takes sequential adjacent frames as input. Features are aggregated in different frames and bounding box regression is conducted with temporal information. More recently, several methods with visual transformers \cite{sun2020transtrack,meinhardt2021trackformer} are also explored in the MOT field and achieve comparable results. For example, \cite{sun2020transtrack} proposes a baseline tracker via a transformer, which takes advantage of the query-key mechanism and introduces a set of learned object queries into the pipeline to enable detecting new-coming objects. \cite{meinhardt2021trackformer} extends the DETR object detector \cite{carion2020end} and achieves a seamless data association between frames in a new tracking-by-attention paradigm by encoder-decoder self-attention mechanisms. However, due to the heavy computational cost, the networks can only take a very limited number of frames as input. As a result, the global motion patterns are still not well utilized.

\section{Method}
As shown in Figure~\ref{fig:Flowchart}, we propose box and tracklet embedding based on GCN to learn both local and global motions in the LGM tracker. Boxes are locally connected to form the tracklets, followed by the tracklet association to further form the tracks. Details are demonstrated in the following sub-sections.

\subsection{Box Motion Embedding}


The box embedding is proposed to associate boxes into tracklets given the box graph with the connection between adjacent frames. Considering a temporal window, we build the box graph based on the adjacency among boxes. Specifically, denote $\boldsymbol{X}^{0} \in \mathbb{R}^{N \times 4}$ as the input boxes with normalized box parameters $x,y,w,h$, where $N$ is the total number of boxes inside the temporal window. Denote $\boldsymbol{A} \in \mathbb{R}^{N \times N}$ as the adjacency matrix, where $\boldsymbol{A}_{ij}=1$ if box $i$ and box $j$ are in the neighboring frames; otherwise set $\boldsymbol{A}_{ij}=0$.

To learn both the structural and temporal relations among detections, inspired from \cite{velivckovic2017graph}, we stack $L$ attention-guided GCN blocks. For the $l$-th GCN layer, the update rule is defined as follows,
\begin{equation}
    \boldsymbol{X}^{l} = \text{ReLU}(\boldsymbol{D}^{l^{-1/2}}\boldsymbol{\hat{A}}^{l}\boldsymbol{D}^{l^{-1/2}}\boldsymbol{X}^{l-1}\boldsymbol{W}^{l})+\boldsymbol{X}^{l-1},
\label{eq:agg}
\end{equation}
where $\boldsymbol{X}^{l-1}$ is the node embedding from the $(l-1)$-th layer, $\boldsymbol{W}^{l}$ is the convolution kernel, $\boldsymbol{\hat{A}}^{l}$ is the refined adjacency matrix and $\boldsymbol{D}^{l}$ is the diagonal node degree matrix with $\boldsymbol{D}_{ii}^{l}=\sum_{j=0} \boldsymbol{\hat{A}}_{ij}^{l}$.

Since most of the connections between adjacent frames among the nodes are from distinct objects, the aggregated information from different objects can have a negative effect on the embedding. As a result, we apply the attention mechanism to refine the adjacency matrix to deal with such an issue as follows,
\begin{equation}
    \boldsymbol{\hat{A}}^{l} = (\boldsymbol{A}+\boldsymbol{I}) \odot \boldsymbol{X}_{att}^{l},
\label{eq:hat_A}
\end{equation}
where $\boldsymbol{X}_{att}^{l}$ is the self-attention feature, $\odot$ represents the elementwise multiplication, and $\boldsymbol{X}_{att}^{l}$ is defined as,
\begin{equation}
    \boldsymbol{X}_{att}^{l} = \sigma(\text{ReLU}( f(\boldsymbol{X}^{l-1}) \boldsymbol{W}_{att,1}^{l})\boldsymbol{W}_{att,2}^{l}),
\label{eq:att}
\end{equation}
where $\boldsymbol{W}_{att,1}$ and $\boldsymbol{W}_{att,2}$ are convolution kernels, $\sigma$ is the Sigmoid activation function and $f$ represents the operation of pairwise self dot product.

\begin{figure*}[!t]
\begin{center}
\includegraphics[width=0.85\linewidth]{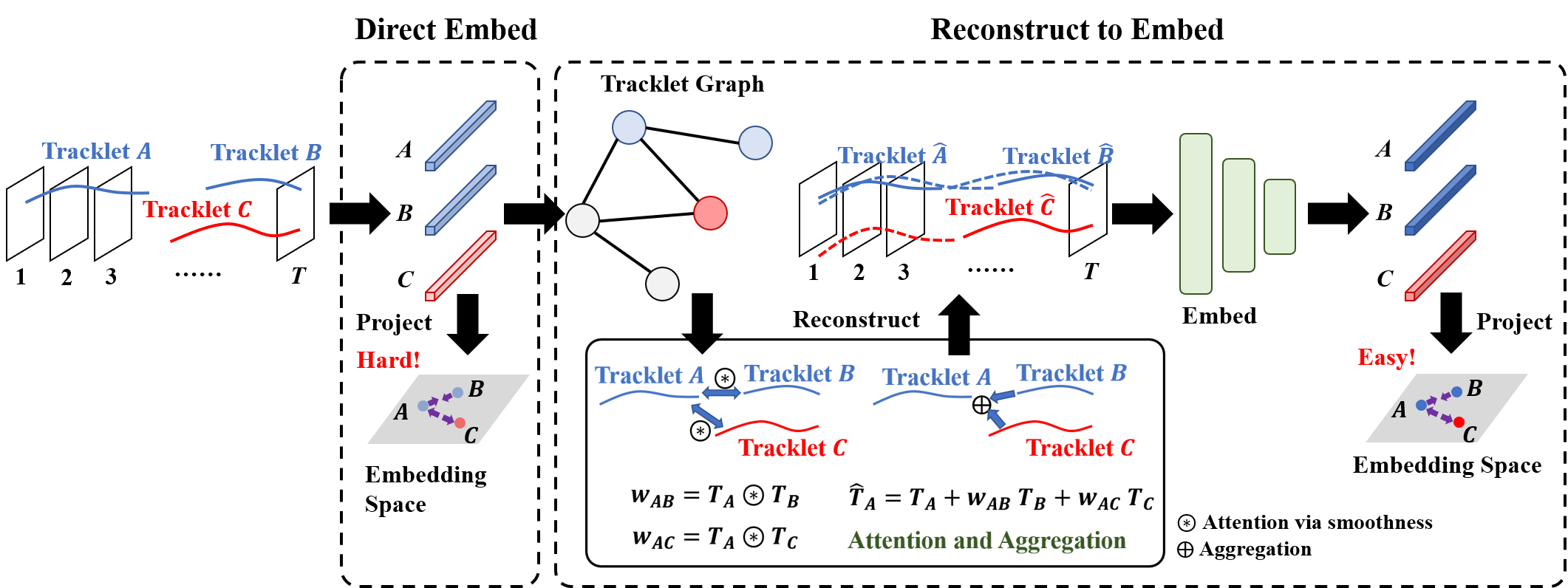}
\end{center}
   \caption{The figure shows the motivation of \textit{reconstruct-to-embed} strategy in the tracklet embedding. The left part is the illustration of the difficulty of the direct tracklet embedding, while the right part shows the proposed indirect tracklet embedding strategy, i.e., \textit{reconstruct-to-embed}. Attention-aggregation mechanism is employed based on the GCN framework in the tracklet reconstruction, followed by the final embedding.}
\label{fig:Recons_emb}
\end{figure*}

We use a combination of triplet loss $\mathcal{L}_{triplet}$ \cite{schroff2015facenet} and binary cross-entropy loss $\mathcal{L}_{xent}$ for training box embedding module. They are defined as follows,
\begin{equation}
\begin{aligned}
    & \mathcal{L}_{triplet} =  \frac{1}{N}\sum_i^N [||\boldsymbol{X}_{i}^{out^a}-\boldsymbol{X}_{i}^{out^p}||_2^2\\& -||\boldsymbol{X}_{i}^{out^a}-\boldsymbol{X}_{i}^{out^n}||_2^2+\alpha]_+,\\
    & \mathcal{L}_{xent} = \frac{1}{N}\sum_{ij}^N \{t_{ij} \log{\boldsymbol{X}_{att,ij}^L} \\
    & +(1-t_{ij}) \log{(1-\boldsymbol{X}_{att,ij}^L)}\},
\label{eq:BLC_trip}
\end{aligned}
\end{equation}
where $[\cdot]_+$ clamps the input value to be non-negative; $\boldsymbol{X}_{i}^{out^a}$, $\boldsymbol{X}_{i}^{out^p}$ and $\boldsymbol{X}_{i}^{out^n}$ represent the box embedding output from the anchor sample, positive sample 
and negative sample, 
respectively; $\alpha$ is the pre-defined margin; $\boldsymbol{X}_{att,ij}^L$ measures the similarity between node $i$ and $j$ in the last GCN layer, while $t_{ij}$ is the binary label indicating the identity between node $i$ and $j$. Therefore, the total loss for the box embedding training is defined as
\begin{equation}
    \mathcal{L}_{box} = \mathcal{L}_{triplet}+\lambda_1\mathcal{L}_{xent}.
\label{eq:loss_BLC}
\end{equation}

\subsection{Tracklet Motion Embedding}
\label{sec:TGA}


The goal of tracklet motion embedding is to explore the global motion patterns among tracklets for the association. However, tracklet embedding is not quite straightforward. 
Since tracklets from the same object exist in different frames and usually have different temporal lengths, it is very challenging to find a latent space to ensure that they share similar feature embeddings. As shown in the left part of Figure~\ref{fig:Recons_emb}, the direct embedding is difficult to measure the similarity among tracklets. To alleviate the challenge in the tracklet embedding, we propose a novel embedding strategy, named as \textit{reconstruct-to-embed}, following an \textit{attention} and \textit{reconstruction} mechanism based on GCN, as shown in the right part of Figure~\ref{fig:Recons_emb}. The motivation behind this is simple. Take tracklet $A$ for example. We calculate the attention from tracklet $B$ and $C$ based on the smoothness of temporal relations. Then based on the attention scores from $B$ and $C$, we aggregate and reconstruct the latent trajectory $\hat{A}$. We use the same reconstruction strategy for $B$ and $C$ to generate $\hat{B}$ and $\hat{C}$. Compared with the original tracklets $A$ and $B$, $\hat{A}$ and $\hat{B}$ have much more similar motion patterns after the reconstruction, which makes the embedding much easier than the situation in the direct embedding. Finally, embeddings are learned with one additional embed-head block with reconstructed tracklets.

The tracklet graph is defined as follows. Considering a temporal window, we denote $\boldsymbol{X}^{0} \in \mathbb{R}^{N \times 4 \times T}$ as the input tracklets with normalized box parameters $x,y,w,h$, where $N$ is the number of tracklets and $T$ is the temporal window size. 
Since tracklets usually have different temporal lengths, we pad zeros along the temporal dimension if the length is smaller than the temporal window $T$. 
We also define tracklet temporal occupancy matrix $\boldsymbol{M}^0 \in \mathbb{R}^{N \times 1 \times T}$ as the mask input, where $\boldsymbol{M}_{it}^{0}=0$ if no box exists for tracklet $i$ in frame $t$ due to missing detection or occlusion; otherwise $\boldsymbol{M}_{it}^{0}=1$. Denote $\boldsymbol{A} \in \mathbb{R}^{N \times N}$ as the adjacency matrix, where $\boldsymbol{A}_{ij}=1$ if tracklet $i$ and tracklet $j$ do not have overlapping frames in the temporal window; otherwise $\boldsymbol{A}_{ij}=0$ following the assumption that temporal overlapping tracklets cannot share the same ID. 

\begin{figure}[t]
\begin{center}
\includegraphics[width=0.85\linewidth]{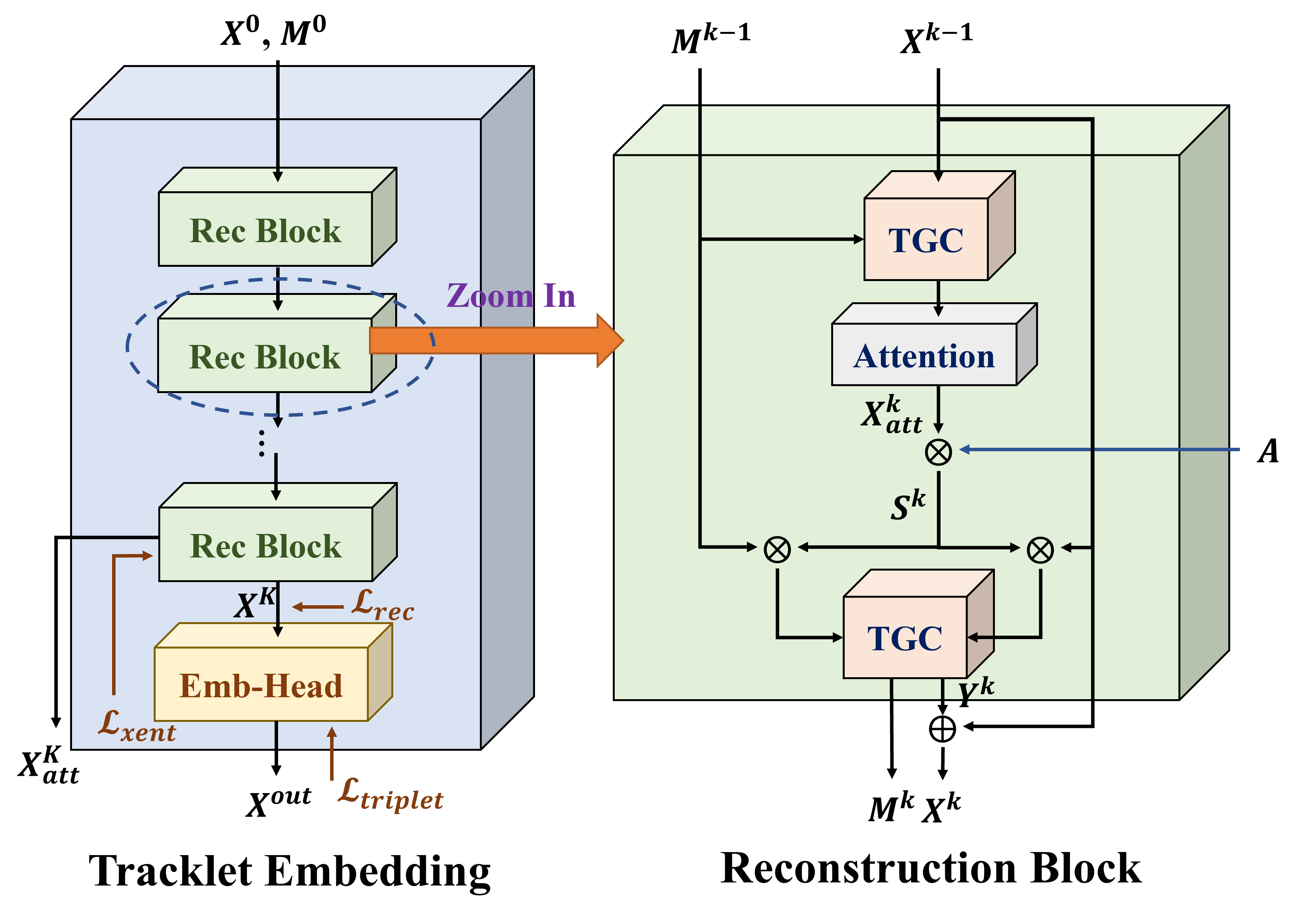}
\end{center}
  \caption{The overview architecture of the tracklet embedding module. The left part shows tracklet embedding with sequentially stacked reconstruction blocks and one embed-head block. Three losses, i.e., triplet loss $\mathcal{L}_{triplet}$, cross-entropy loss $\mathcal{L}_{xent}$ and reconstruction loss $\mathcal{L}_{rec}$, are combined in the training. The right part shows each basic reconstruction block based on GCN and temporal gated convolution (TGC) modules for temporal information extraction. }
\label{fig:TGA}
\end{figure}

The architecture of tracklet embedding is shown in Figure~\ref{fig:TGA}. We stack $K$ reconstruction blocks to measure the long-term relations in both structural and temporal dimensions for reconstruction, and then followed by one embed-head block for the final embedding. 
As shown in Figure~\ref{fig:TGA}, each reconstruction block updates the feature maps and soft masks based on the attention and temporal gated convolution (TGC) module. 
The goal of attention is to measure the similarities among tracklets. Good tracklet embeddings should have high similarities if they are from the same object. However, simple operations cannot well represent the similarities. If we take cosine similarity measure for example, the zero-padded motion features of two tracklets from the same object would be always zero due to the fact that two tracklets from the same object do not have overlapping frames. 
Inspired by the TGC module in the inpainting task \cite{dauphin2017language,yu2019free,gu2020exploring}, where missing values can be reconstructed with input masks, we can use TGC with the occupancy masks for tracklet extrapolation. Based on the extrapolated tracklets, the similarities are easier to measure and the temporal consistency among tracklets from the same object can be exploited. 
Specifically, the attention/similarity map of the $k$-th reconstruction block is calculated as follows,
\begin{equation}
\begin{aligned}
    & \tilde{\boldsymbol{X}}^{k}, \tilde{\boldsymbol{M}}^{k} = g(\boldsymbol{X}^{k-1}, \boldsymbol{M}^{k-1}), \\
    & \boldsymbol{X}_{att}^{k} = \sigma(\text{ReLU}( f(\tilde{\boldsymbol{X}}^{k}) \boldsymbol{W}_{att,1}^{k})\boldsymbol{W}_{att,2}^{k}),
\label{eq:tgc_tmp}
\end{aligned}
\end{equation}
where $\boldsymbol{X}^{k-1}$ and $\boldsymbol{M}^{k-1}$ are feature maps and masks from the previous block, $g$ is the TGC module and $f$ is the pairwise self dot product operation. Here, we calculate attention maps after the extrapolation of tracklets based on TGC module. 
The TGC module has multiple stacked TGC layers. To be specific, the feature map and mask of each TGC layer are updated as follows,
\begin{equation}
\begin{aligned}
    & \boldsymbol{M}^{j+1} = \sigma (\boldsymbol{M}^{j} \ast \boldsymbol{W}_{M}^j), \\
    & \boldsymbol{Y}^{j+1} = \text{ReLU}(\boldsymbol{Y}^{j} \ast \boldsymbol{W}_{Y}^j) \odot \boldsymbol{M}^{j+1}+\boldsymbol{Y}^{j},
\label{eq:gated_conv}
\end{aligned}
\end{equation}
where $\ast$ represents the convolution operation, $\odot$ represents the dot product, $\boldsymbol{M}^{j}$ is the soft mask, $\boldsymbol{Y}^{j}$ is the feature mask, $\boldsymbol{W}_{M}^j$ and $\boldsymbol{W}_{Y}^j$ are the convolution kernel weights for the mask and feature maps, respectively.




Then we aggregate the attention maps based on the graph structure. The aggregation output $\boldsymbol{S}^{k}$ is represented as 
\begin{equation}
    \boldsymbol{S}^{k} = \boldsymbol{D}^{k^{-1/2}}\boldsymbol{\hat{A}}^{k}\boldsymbol{D}^{k^{-1/2}},
\label{eq:agreg}
\end{equation}
where $\boldsymbol{\hat{A}}^{k}$ is with the same definition as Eq.~(\ref{eq:hat_A}), which re-calculates the attention maps based on the adjacency matrix of the graph structure.

We aggregate the attention map $\boldsymbol{S}^{k}$ with $\boldsymbol{X}^{k-1}$ and $\boldsymbol{M}^{k-1}$, followed by a second TGC module to obtain the output of the $k$-th reconstruction block as follows, 
\begin{equation}
\begin{aligned}
    & \boldsymbol{Y}^{k}, \boldsymbol{M}^{k} = g(\boldsymbol{S}^{k} \boldsymbol{X}^{k-1}, \boldsymbol{S}^{k} \boldsymbol{M}^{k-1}), \\
    & \boldsymbol{X}^{k} = \boldsymbol{X}^{k-1}+\boldsymbol{Y}^{k}.
\label{eq:TGC_B}
\end{aligned}
\end{equation}
Simply put, the second TGC module plays a role as a further transformation of node embeddings.

To obtain the final tracklet embedding, a feed-forward transformation with two dense layers is included in the embed-head block. The feed-forward transformation is defined as follows,
\begin{equation}
    \boldsymbol{X}^{out} = \text{Norm}_{l_2}(\text{ReLU}( f(\boldsymbol{X}^{K}) \boldsymbol{W}_{1})\boldsymbol{W}_{2}),
\label{eq:fw}
\end{equation}
where $\text{Norm}_{l_2}$ is the $l_2$ normalization, $\boldsymbol{W}_{1}$ and $\boldsymbol{W}_{2}$ are the weights of the two dense layers.

We also adopt the triplet loss $\mathcal{L}_{triplet}$ and binary cross-entropy loss $\mathcal{L}_{xent}$ based on the output $\boldsymbol{X}^{out}$ for tracklet embedding training with similar equations to Eq.~(\ref{eq:BLC_trip}).
In addition, a reconstruction loss $\mathcal{L}_{rec}$ is employed in the tracklet embedding. To be specific, the reconstruction loss is defined as follows,
\begin{equation}
    \mathcal{L}_{rec} = ||\boldsymbol{M}^*(\boldsymbol{X}^K-\boldsymbol{X}^*)||_2,
\label{eq:TGA_rec}
\end{equation}
where $\boldsymbol{X}^*$ is the ground truth track and $\boldsymbol{M}^*$ is the ground truth occupancy mask.
Finally, the total loss for tracklet embedding training is defined as follows,
\begin{equation}
    \mathcal{L}_{tracklet} =\mathcal{L}_{triplet}+\lambda_2\mathcal{L}_{xent}+\lambda_3\mathcal{L}_{rec}.
\label{eq:loss}
\end{equation}

\subsection{Inference}
After training, the association is conducted based on the box and tracklet embedding in the inference stage. The temporal sliding window procedure is adopted with 50\% overlapping frames. Within each temporal window, boxes are associated based on the embedding distances. Then the tracklets are associated with the bottom-up greedy method. 50\% overlap is used to ensure the new boxes and tracklets can be matched to existing tracked objects.

\section{Experiments}

\begin{table*}[!t]
\begin{center}
\begin{tabular}{|l|c|c|c|c|c|c|c|c|}
\hline
Method & HOTA (\%) $\uparrow$ & AssA (\%) $\uparrow$ & MOTA (\%) $\uparrow$ & MT (\%) $\uparrow$ & ML (\%) $\downarrow$ & IDS $\downarrow$ & FRAG $\downarrow$ \\
\hline\hline
$^{\dagger \star}$JRMOT \cite{shenoi2020jrmot} & 69.6 & 66.9 & \color{red}85.1 & 70.9 & 4.6 & 271 & 273\\
$^{\dagger \star}$AB3DMOT \cite{weng20203d} & 69.8 & 69.1 & 83.5 & 67.1 & 11.4 & \color{red}126 & \color{red}254\\
$^{\dagger \star}$MOTSFusion \cite{luiten2020track} & 68.7 & 66.2 & 84.2 & 72.8 & \color{red}2.9 & 415 & 569\\
$^{\dagger \star}$mono3DT \cite{hu2019joint} & \color{red}73.2 & \color{red}74.2 & 84.3 & \color{red}73.1 & \color{red}2.9 & 379 & 573 \\
\hline
$^{\star}$MASS \cite{karunasekera2019multiple} & 68.3 & 64.5 & 84.6 & 74.0 & 2.9 & 353 & 516\\
$^{\star}$TuSimple \cite{choi2015near} & 71.6 & 71.1 & 86.3 & 71.1 & 6.9 & 292 & \color{blue}220 \\
$^{\star}$SMAT \cite{gonzalez2020smat} & 71.9 & \color{blue}72.1 & 83.6 & 62.8 & 6.0 & \color{blue}198 & 294 \\
$^{\star}$CenterTrack \cite{zhou2020tracking} & \color{blue}73.0 & 71.2 & \color{blue}88.8 & \color{blue}82.2 & \color{blue}2.5 & 254 & 227 \\
\hline
DCO-X \cite{milan2013detection} & 46.5 & 38.7 & 66.2 & 38.3 & 14.5 & 955 & 708\\
SCEA \cite{hong2016online} & 56.1 & 52.2 & 74.9 & 53.7 & 12.3 & \textbf{324} & 317 \\
MCMOT-CPD \cite{lee2016multi} & 56.6 & 50.6 & 78.0 & 52.5 & 12.5 & 475 & 309\\
\textbf{LGM (ours)} & \textbf{73.1} & \textbf{72.3} & \textbf{87.6} & \textbf{85.1} & \textbf{2.5} & 448 & \textbf{164} \\
\hline
\end{tabular}
\end{center}
\caption{Result on KITTI-car tracking testing set. From top to bottom, we divide SOTA methods into three categories (\textit{3D trackers}, \textit{2D trackers}, \textit{mere motion 2D trackers}) according to different input information. 
$^{\dagger}$ represents 3D tracking methods and $^{\star}$ represents trackers using appearance information. The best result for each part is shown in red, blue and bold, respectively. }
\label{tab:KITTI}
\end{table*}

\begin{table*}[!t]
\begin{center}
\begin{tabular}{|l|c|c|c|c|c|c|c|}
\hline
Method & PR-MOTA (\%) $\uparrow$ & PR-MOTP (\%) $\uparrow$ & PR-MT (\%) $\uparrow$ & PR-ML (\%) $\downarrow$ & PR-IDS $\downarrow$ & PR-FRAG $\downarrow$ \\
\hline\hline
IHTLS \cite{dicle2013way} & 11.1 & 36.8 & 13.8 & 19.9 & 953.6 & 3556.9 \\
H2T \cite{wen2014multiple} & 12.4 & 35.7 & 14.8 & 19.4 & 852.2 & 1117.2 \\
CMOT \cite{bae2014robust} & 12.6 & 36.1 & 16.1 & 18.6 & \textbf{285.3} & 1516.8 \\
GOG \cite{pirsiavash2011globally} & 14.2 & 37.0 & 13.9 & 19.9 & 3334.6 & 3172.4 \\
IOUT \cite{1517Bochinski2017} & 16.1 & \textbf{37.0} & 14.8 & 19.7 & 2308.1 & 3250.4 \\
V-IOUT \cite{1547Bochinski2018} & 17.7 & 36.4 & \textbf{17.4} & 18.8 & 363.8 & 1123.5 \\
FAMNet \cite{chu2019famnet} & 19.8 & 36.7 & 17.1 & 18.2 & 617.4 & \textbf{970.2} \\
\textbf{LGM (ours)} & \textbf{22.5} & 35.2 & 15.5 & \textbf{10.1} & 1563.5 & 3186.8 \\
\hline
\end{tabular}
\end{center}
\caption{Result on UA-Detrac testing set. The best performance is shown in bold type.}
\label{tab:UA-Detrac}
\end{table*}

\subsection{Datasets}
Two vehicle tracking benchmark datasets, i.e., KITTI \cite{Geiger2012CVPR} and UA-Detrac \cite{wen2020ua}, are used for validation.

\textbf{KITTI.} 
The KITTI car tracking benchmark consists of 21 training sequences and 29 testing sequences. Videos are captured at 10 FPS and contain large inter-frame motions. We only evaluate the tracking performance on the \textit{car} category.

\textbf{UA-Detrac.} 
The UA-DETRAC is a large-scale tracking dataset for vehicles. It comprises 100 videos that record around 10 hours of vehicle traffic. The recording is made in 24 different locations, and it includes a wide variety of common vehicle types and traffic conditions. 
Overall, the dataset contains about 140k video frames, 8,250 vehicles, and 1,210k bounding boxes. 

\subsection{Implementation Details}
For the architecture of the proposed LGM tracker, we stack $L=8$ GCN blocks and $K=4$ reconstruction blocks in the box and the tracklet embedding modules, respectively. For the TGC module in the tracklet embedding, we have 6 basic TGC layers. We use 17 frames and 65 frames as the temporal window for the box and tracklet embedding modules, respectively. The final embedding dimension for the box and tracklet is set to $D=128$. The margin $\alpha$ for calculating the triplet loss is set to 0.2. For the loss combination, we simply set all $\lambda$s in Eq.~(\ref{eq:loss_BLC}) and Eq.~(\ref{eq:loss}) to 1.

We use detection results from CenterNet \cite{zhou2019objects,zhou2020tracking} and CompACT \cite{cai2015learning} as our input boxes for KITTI and UA-Detrac datasets, respectively. 
Both the box and tracklet embedding modules are trained with Adam optimizer \cite{kingma2014adam} with an initial learning rate of 1e-3. We use a cosine annealing learning rate scheduler for the learning rate decay. The maximum step is set to 200000. 

Data augmentation strategy is adopted for training the LGM tracker. For the box embedding module, the input boxes are pre-processed with random horizontal flips, randomly jittered sizes and positions. We also randomly add boxes as false positives and remove some ground truth boxes as false negatives. A similar augmentation strategy is used for training the tracklet embedding module. Besides that, we also randomly split the ground truth tracks into pieces of tracklets for the augmentation.

\subsection{Evaluation Metrics}
We use the default metrics defined by the benchmark datasets for the evaluation. The metrics include Higher Order
Tracking Accuracy (HOTA) \cite{luiten2020hota}, Association Accuracy (AssA), Multiple Object Tracking Accuracy (MOTA) \cite{milan2016mot16}, ID F1 score (IDF1), Multiple Object Tracking Precision (MOTP) \cite{milan2016mot16}, the number of ID Switches (IDS), the percentage of Mostly Tracked targets (MT), the percentage of Mostly Lost targets (ML) and the total number of times a trajectory is Fragmented (FRAG). HOTA and AssA are newly defined in \cite{luiten2020hota} and adopted as the main evaluation metrics for MOT in the KITTI benchmark dataset. For the UA-Detrac dataset, the metrics with PR-curve integrated are used, as defined in \cite{wen2020ua}.

\subsection{Main Results on Benchmark Datasets}

\textbf{KITTI-Car Tracking Benchmark.}
The tracking result on the KITTI-car testing dataset is shown in Table~\ref{tab:KITTI}. 
From top to bottom, we divide SOTA methods into three categories i.e., $^{\dagger}$\textit{3D trackers}, $^{\star}$\textit{2D trackers}, and \textit{mere motion 2D trackers}, according to different input information. 

We achieve the best performance among the mere motion trackers, and the proposed LGM tracker is also very competitive to other SOTA methods. The result demonstrates the effectiveness of the proposed LGM tracker based on mere motion information.

\textbf{UA-Detrac Benchmark.}
The tracking result on the UA-Detrac testing dataset with CompACT detections is shown in Table~\ref{tab:UA-Detrac}. We can see that the proposed LGM tracker outperforms most SOTA methods, including methods that use both appearance and motion information. 

\subsection{Qualitative Results}

\begin{figure*}[t]
\begin{center}
\includegraphics[width=0.95\linewidth]{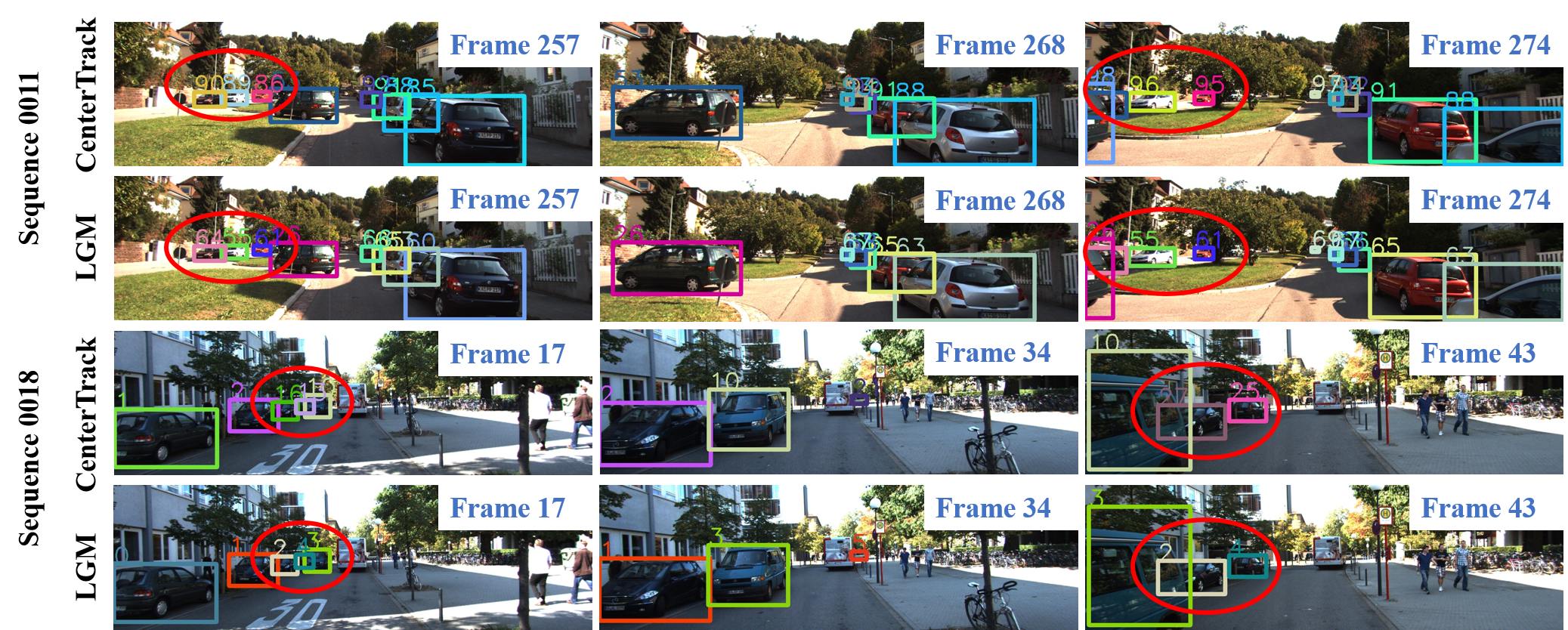}
\end{center}
   \caption{Examples of occlusion handling. Each row shows three frames from the same sequence. Each tracked vehicle is represented in a unique color. The number on the bounding box is the tracking ID.}
\label{fig:test}
\end{figure*}

\textbf{Occlusion Handling.}
We show some qualitative examples of the proposed LGM tracker against CenterTrack \cite{zhou2020tracking} about occlusion handling on the KITTI-car testing dataset in Figure~\ref{fig:test}. Each color represents a distinct tracked object with the tracking ID on the top of the bounding box. The first two rows are results of CenterTrack and LGM tracker from sequence 0011, respectively. For the LGM tracker, three cars with a red circle drawn on the figure are occluded by a car in frame 268 and then associated to the correct labels in frame 274 after they reappear, while the CenterTrack fails in the association. The last two rows show another example from sequence 0018, where two cars in the red circle are correctly associated in frame 43 for the LGM tracker, while CenterTrack fails again. These two examples demonstrate the robustness of the proposed tracker's occlusion handling strategy.

\begin{figure}[t]
\begin{center}
\includegraphics[width=0.9\linewidth]{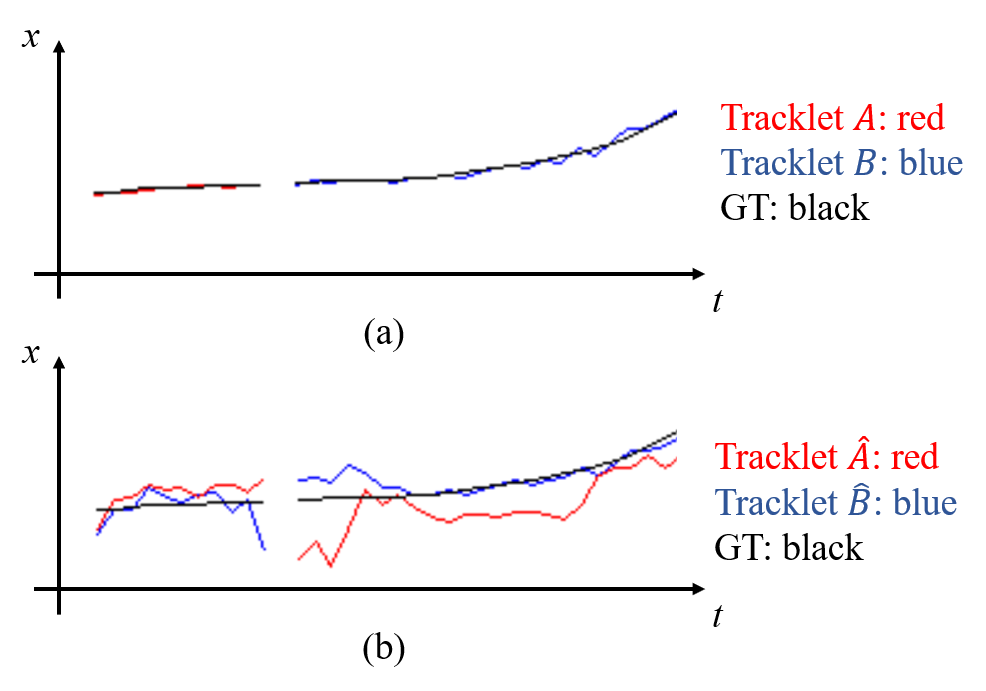}
\end{center}
   \caption{Visualization example of tracklet reconstruction. (a) shows two input tracklets, $A$ and $B$. Both $A$ and $B$ are from the same object with the ground truth trajectory shown in black. (b) shows the reconstructed $\hat{A}$ and $\hat{B}$ after the last reconstruction block.}
\label{fig:rec}
\end{figure}

\textbf{Reconstruction Analysis on the Tracklet.}
To further illustrate the \textit{reconstruct-to-embed} strategy in the tracklet embedding module,
we also provide one visualization example, as shown in Figure~\ref{fig:rec}. For Figure~\ref{fig:rec} (a), we plot two tracklets $A$ and $B$ from the same object in red and blue colors, respectively. The ground truth trajectory is displayed in black. Figure~\ref{fig:rec} (b) shows the reconstructed tracks, $\hat{A}$ and $\hat{B}$, after the last reconstruction block. Based on the visualization, it is obvious that $\hat{A}$ and $\hat{B}$ have much higher similarities than $A$ and $B$, which makes the embedding much easier.

\subsection{Ablation Study}

\begin{table}[!t]
\begin{center}
\begin{tabular}{|l|c|c|c|}
\hline
Loss Combination & HOTA & AssA & MOTA \\
\hline\hline
$\mathcal{L}_{triplet}$ & 75.7 & 75.0 & 88.0 \\
$\mathcal{L}_{triplet}+\mathcal{L}_{xent}$ & \textbf{76.9} & \textbf{77.2} & \textbf{89.0} \\
\hline
\end{tabular}
\end{center}
\caption{Results of different loss combinations for box embedding module on the KITTI-car tracking validation set.}
\label{tab:box_module}
\end{table}

\begin{table}[!t]
\begin{center}
\begin{tabular}{|l|c|c|c|}
\hline
Loss Combination & HOTA & AssA & MOTA \\
\hline\hline
$\mathcal{L}_{triplet}$ & - & - & - \\
$\mathcal{L}_{triplet}+\mathcal{L}_{xent}$ & - & - & - \\
$\mathcal{L}_{triplet}+\mathcal{L}_{rec}$ & 76.3 & 76.5 & 88.2 \\
$\mathcal{L}_{triplet}+\mathcal{L}_{xent}+\mathcal{L}_{rec}$ & \textbf{76.9} & \textbf{77.2} & \textbf{89.0}\\
\hline
\end{tabular}
\end{center}
\caption{Results of different loss combinations for tracklet embedding module on the KITTI-car tracking validation set.
Results for the first two combinations are not available since the training does not converge.}
\label{tab:loss}
\end{table}

\textbf{Study of Loss Combination for Box Embedding.} 
The study of loss combination for box embedding on the KITTI-car validation set with the same data split defined from \cite{zhou2020tracking} is shown in Table~\ref{tab:box_module}, where the first row only uses triplet loss in the training while the second row uses both triplet loss and binary cross-entropy loss. The same tracklet embedding module is used for both cases. We can see that with a combination of the two losses, the performance is better.

\textbf{Study of Loss Combination for Tracklet Embedding.}
To better understand the importance of each loss term in the tracklet embedding module, we try four different loss combinations in the training and then evaluate the KITTI-car validation set. For the first two trials, with standalone triplet loss $\mathcal{L}_{triplet}$ or with a combination of triplet loss and binary cross-entropy loss $\mathcal{L}_{triplet}+\mathcal{L}_{xent}$, the model fails to converge. This demonstrates the necessity of the reconstruction loss in the training and it further proves the effectiveness of \textit{reconstruct-to-embed} strategy illustrated in Figure~\ref{fig:Recons_emb}. As shown in the last two rows of Table~\ref{tab:loss}, the performance improves when introducing the binary cross-entropy loss $\mathcal{L}_{xent}$, which shows the importance of the supervision on the attention mechanism.

\begin{table}[!t]
\begin{center}
\begin{tabular}{|l|c|c|c|}
\hline
Module & HOTA & AssA & MOTA \\
\hline\hline
Box & 75.2 & 74.3 & 88.0\\
Box+Tracklet & \textbf{76.9} & \textbf{77.2} & \textbf{89.0}\\
\hline
\end{tabular}
\end{center}
\caption{Results of different modules on the KITTI-car tracking validation set.}
\label{tab:module}
\end{table}

\textbf{Analysis on Functionalities of Box and Tracklet Embedding.}
We also test the functionalities of the box and tracklet embedding modules on the KITTI-car validation set. The result is shown in Table~\ref{tab:module}. We can see that with box embedding alone we can achieve tolerable results, yet the tracklet association is not exploited. From the second row of the table, there is further improvement in the tracking performance with the tracklet embedding module added. This example demonstrates the importance of both box and tracklet embedding.

\subsection{Generalization to Pedestrian Tracking}

Although pedestrian tracking is not the focus of this paper, we still report the results for pedestrian tracking on the MOT17 testing set, as shown in Table~\ref{tab:ped}, where the top three methods are widely used SOTA methods with both appearance and other clues while the bottom two only use motion clues for tracking. Pedestrian tracking is more challenging than vehicle tracking for motion-based trackers since the motion consistency assumption is not always the truth. Due to such challenges, we can still achieve comparable results using the proposed motion tracker, demonstrating the generalization ability to pedestrian tracking.

\begin{table}[!t]
\begin{center}
\begin{tabular}{|l|c|c|c|}
\hline
Method & MOTA & IDF1 & MOTP \\
\hline\hline
Tracktor++ \cite{bergmann2019tracking} & 56.3 & 55.1 & 78.8 \\
TrctrD17 \cite{xu2020train} & 53.7 & 53.8 & 77.2 \\
CenterTrack \cite{zhou2020tracking} & \noindent\color{blue}61.5 & \noindent\color{blue}59.6 & \noindent\color{blue}78.9\\
\hline
IOU Tracker & 45.5 & 39.4 & 76.9 \\
\textbf{LGM (ours)} & \textbf{56.0} & \textbf{55.6} & \textbf{78.0} \\
\hline
\end{tabular}
\end{center}
\caption{Pedestrian tracking result on MOT17 dataset, where top three methods use appearance information while the bottom two methods only employ motion features.}
\label{tab:ped}
\end{table}

\section{Conclusion}
In this paper, we propose a novel tracker with motion consistency without looking at the appearance for the vehicle tracking task. Two modules, box and tracklet embedding, are designed to model both local and global motion information based on deep convolutional networks. We evaluate the proposed method on two vehicle tracking datasets, i.e., KITTI-car tracking benchmark and UA-Detrac benchmark and achieve competitive results with mere motion information. We also visualize the tracking results and the reconstructed trackets in the tracklet embedding module. This further proves the effectiveness of the proposed \textit{reconstruct-to-embed} strategy. Several ablation studies are conducted to show the importance of each module and the losses in the model training. 
In future work, we plan to to incorporate appearance information in both the box and tracklet embedding modules for further improvement.

\noindent \textbf{Acknowledgement}
This work is supported by National Natural Science Foundation of China (U20B2066, 61976186),  the Major Scientific Research Project of Zhejiang Lab (No. 2019KD0AC01), the Fundamental Research Funds for the Central Universities, Alibaba-Zhejiang University Joint Research Institute of Frontier Technologies, Guangdong Provincial Characteristic Innovation Natural Science Projects of Colleges and Universities, China (2019ktscx110) and Guangdong Provincial Special Funding projects for Introducing Innovation Team and Industry University Research Cooperation, China (2019C002001).

{\small
\bibliographystyle{ieee_fullname}
\bibliography{egbib}
}

\end{document}